\documentclass{article}

\PassOptionsToPackage{square, numbers}{natbib}

\usepackage{amsmath}
\usepackage[preprint]{dfkd_arxiv}

\usepackage[utf8]{inputenc} 
\usepackage[T1]{fontenc} 
\usepackage{hyperref} 
\usepackage{url} 
\usepackage{booktabs} 
\usepackage{amsfonts} 
\usepackage{nicefrac} 
\usepackage{microtype} 
\usepackage{xcolor} 
\usepackage{graphicx} 
\usepackage{bm}
\usepackage{multirow}
\usepackage{makecell}
\usepackage{wrapfig}
\usepackage{floatrow}
\usepackage[marginal]{footmisc}

\newfloatcommand{capbtabbox}{table}[][\FBwidth]

\newcommand{\eg}{\emph{e.g.}}
\newcommand{\ie}{\emph{i.e.}}
\newcommand{\bx}{\bm{x}}
\newcommand{\by}{\bm{y}}
\newcommand{\bp}{\bm{p}}
\newcommand{\bt}{\bm{t}}
\newcommand{\bs}{\bm{s}}
\newcommand{\dQ}{\mathbb {Q}}
\newcommand{\dP}{\mathbb {P}}
\newcommand{\lsp}{\mathcal L_{\rm spec}}
\newcommand{\lce}{\mathcal L_{\rm CE}}

\newcommand{\ldiversity}{\mathcal{L}_{\rm divs}}
\newcommand{\ldivergence}{\mathcal{L}_{\rm divg}}
\newcommand{\ltt}{\mathcal{L}_{\rm ttgen}}
\newcommand{\lkd}{\mathcal{L}_{\rm KD}}
\newcommand{\lhid}{\mathcal{L}_{\rm hidd}}

\newcommand{\zheyuan}[1]{{\textcolor{red}{#1}}}

\title{Data-Free Distillation of Language Model by Text-to-Text Transfer}

\author{Zheyuan Bai$^{*}$, ~~Xinduo Liu\thanks{Equal contribution. $^\dagger$ Corresponding author.}, ~~Hailin Hu, ~~Tianyu Guo, ~~Qinghua Zhang, ~~Yunhe Wang$^{\dagger}$ \\
     Huawei Noah's Ark Lab \\
	 \texttt{\{baizheyuan, liuxinduo, yunhe.wang\}@huawei.com} \\
}

\begin{document}

\maketitle


\begin{abstract}
Data-Free Knowledge Distillation (DFKD) plays a vital role in compressing the model when original training data is unavailable. Previous works for DFKD in NLP mainly focus on distilling encoder-only structures like BERT on classification tasks, which overlook the notable progress of generative language modeling. In this work, we propose a novel DFKD framework, namely DFKD-T$^{3}$, where the pretrained generative language model can also serve as a controllable data generator for model compression. This novel framework DFKD-T$^{3}$ leads to an end-to-end learnable text-to-text framework to transform the general domain corpus to compression-friendly task data, targeting to improve both the \textit{specificity} and \textit{diversity}. Extensive experiments show that our method can boost the distillation performance in various downstream tasks such as sentiment analysis, linguistic acceptability, and information extraction. Furthermore, we show that the generated texts can be directly used for distilling other language models and outperform the SOTA methods, making our method more appealing in a general DFKD setting. Our code is available at https://gitee.com/mindspore/models/tree/master/research/nlp/DFKD\_T3.
\end{abstract}

\section{Introduction}

\begin{figure*}[htbp]
\begin{center}
\includegraphics[width=12cm] {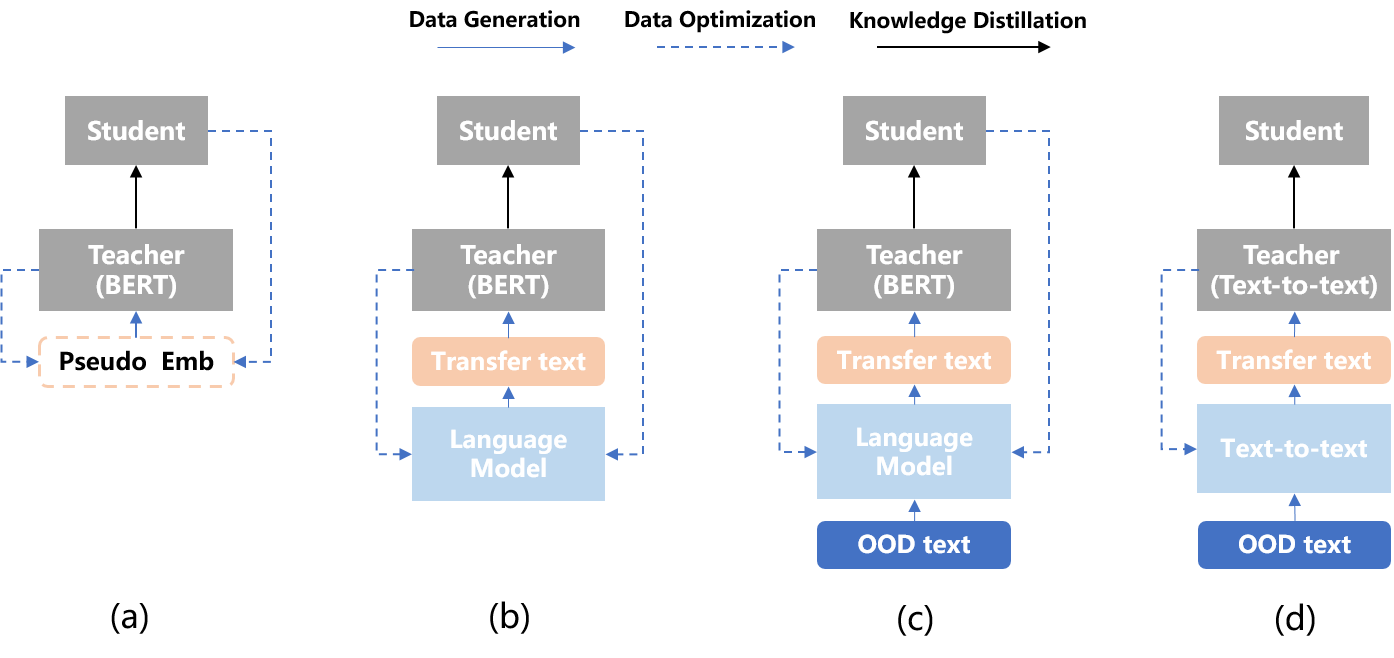}
\end{center}
\caption{Architecture comparison of different DFKD methods in NLP. (a)-(c) highlights the major design principles for encoder-based DFKD, which involves pseudo embedding generation (\eg~\cite{DBLP:conf/emnlp/MaSFCJL20}), language model generation(\eg~\cite{DBLP:conf/ijcai/MaWF0022}) or OOD data optimization(\eg~\cite{DBLP:conf/emnlp/RashidLGR21}). In our framework~(d), the text-to-text scheme enables the generation of training data from large-scale public corpus with a generative language model. The blue dash line in the figure indicates which model provides information to optimize the training data generated for distillation.}
\label{fig:comparison}
\end{figure*}

Knowledge distillation is an essential method in the compression of deep learning models. It has shown promising performance in various tasks of computer vision~\cite{DBLP:journals/corr/HintonVD15} and natural language processing~\cite{DBLP:journals/corr/abs-1910-01108, DBLP:conf/emnlp/JiaoYSJCL0L20}. However, the success of the traditional distillation method relies heavily on the availability of original training data, which is not guaranteed in real-world settings due to privacy, transmission, or regulatory constraints.

To tackle the difficulty of data availability, data-free knowledge distillation (DFKD) has become an attractive direction of exploration. Originating from the computer vision community, this method leverages a generative model \eg~GAN~\cite{DBLP:conf/iccv/ChenW0YLSXX019} to facilitate the training process. Correspondingly, in natural language tasks, some pioneering works have also explored the application of DFKD and shown promising results on benchmark tasks such as text classification~\cite{DBLP:conf/emnlp/MaSFCJL20, DBLP:conf/emnlp/RashidLGR21, DBLP:conf/ijcai/MaWF0022}.

Despite this progress, the core difficulty of data-free knowledge distillation remains the generation of compression-friendly training data. Currently, most methods in DFKD for language models rely on model inversion~\cite{Mahendran2014UnderstandingDI} and adversarial training~\cite{DBLP:conf/emnlp/MaSFCJL20, DBLP:conf/emnlp/RashidLGR21, DBLP:conf/ijcai/MaWF0022}, which is not guaranteed to generate valid text for compression. Also, people have explored guiding the generation by topic prompt from the decoder-only language model~\cite{DBLP:conf/ijcai/MaWF0022}, which is limited by the capability of the original decoder and cannot obtain knowledge from external corpus. Moreover, all previous works focus on the DFKD of the encoder-only language model (\ie~BERT) on the classification task. This limits their application in other NLP tasks, especially the generation task that is developing very fast due to the recent arising of large language model~\cite{ouyang2022training}.

On the other hand, the past few years have witnessed the emergence of generative language models~\cite{Radford2018ImprovingLU, Brown2020LanguageMA}, whose performance has been validated on various downstream tasks from natural language understanding (\eg~classification, information extraction) to generation. In particular, text-to-text~\cite{DBLP:journals/jmlr/RaffelSRLNMZLL20} has emerged as a common paradigm of applying language models, and it turns out to require less tunable parameters (\ie~prompt tuning) to guide the generation of output to address specific downstream tasks.

Following this motivation, in this work, we design a transfer text generator that can be trained in an end-to-end manner to deliver a particular distillation corpus for a downstream task (Figure~\ref{fig:comparison}). In particular, a generative language model is trained to improve both task specificity and sample diversity for distillation from the general domain corpus. In contrast to previous works, the introduction of the trainable transfer text generator naturally extends the DFKD setting to generative language models and also more tasks, such as entity extraction, in a unified optimization framework. Extensive experiments validate the effectiveness of our method.

\section{Prelimilaries: Data-driven and data-free knowledge distillation}

Distillation~\cite{DBLP:journals/corr/HintonVD15} is a widely used method to transfer knowledge from a larger teacher model to a smaller student model. Although ground truths are not necessarily available, the student model can extract rich knowledge by imitating the distribution of teachers' outputs. At a high level, this objective can be expressed as:
\begin{equation} \label{eq_00}
\begin{split}
\ldivergence = \mathcal{D}(\bm{y}_S, \bm{y}_T),
\end{split}
\end{equation}
where $\mathcal{D}$ is the divergence function, and $\bm{y}_S$ and $\bm{y}_T$ denotes the outputs from either the teachder and student models, respectively. In language model, apart from the final output, the distillation process also involves knowledge transfer using other model layers~\cite{DBLP:conf/emnlp/JiaoYSJCL0L20}. For instance, to distill the hidden states, learnable linear transformation matrices can be introduced to transform the hidden states of the student model into the same space to match the teacher model, \ie,
\begin{equation} \label{eq_12}
\begin{split}
\lhid = \|(\bm h_S \bm{W}_h, \bm h_T)\|_2,
\end{split}
\end{equation}
where $\bm h_S, \bm h_T$ indicate the normalized hidden states of student and teacher networks, respectively. Divergence can also be calculated with embedding or attention matrices in the KD process.

Specifically, when the training data for distillation is available, the KD process is \textit{data-driven}, which has proved to be effective for language models~\cite{DBLP:conf/emnlp/SunCGL19, DBLP:journals/corr/abs-1910-01108, DBLP:conf/emnlp/JiaoYSJCL0L20, DBLP:conf/nips/HouHSJCL20,DBLP:conf/acl/XiaZC22}. In contrast, when the training dataset that is specific for the distilled task is unavailable for some reasons, \eg~transmission limitations and privacy issues, the KD process is constrained in a \textit{data-free} setting, and the distillation performance is primarily hindered. In this case, the core issue to be addressed is to generate proper training data, basically through the modification of general domain corpus~\cite{DBLP:conf/emnlp/RashidLGR21} or applying generative models~\cite{DBLP:conf/emnlp/MaSFCJL20}.


\section{Related Work}
\subsection{Data-free knowledge distillation in NLP}
Several methods have been proposed for the data-free setting in natural language processing. \cite{DBLP:conf/emnlp/MaSFCJL20} generates pseudo-embeddings to address the discrete problem of tokens. \cite{DBLP:conf/emnlp/RashidLGR21} combines out-of-domain data and adversarial training to learn the teacher's output distribution and adopts GumbelSoftmax to pass the gradient to the generator, which has the gap between out-of-domain data and required sentences. \cite{DBLP:conf/ijcai/MaWF0022} leverages a pretrained auto-regressive language model and a topic prompter to control data synthesis. However, these methods only focus on distilling encoder-like networks like BERT~\cite{DBLP:conf/naacl/DevlinCLT19}, and mainly on classification tasks.

Unlike the aforementioned methods, our data-free distillation scheme tries to transfer knowledge between encoder-decoder architectures. In addition to performing well on classification tasks, we also achieved good results on extraction tasks, proving our framework's feasibility for generative model compression.

\subsection{Prompt-tuning}

Unlike traditional supervised learning, which trains a model to take in an
input $x$ and predict an output $y$ as $P(y|x)$, prompt tuning retrieves knowledge without tuning the whole parameter set of the large language model. Instead, it tunes a small number of parameters to change the "prefix" of the input and guide the output generation. For instance, some methods leverage hard prompt(template) for retrieving knowledge from language models, such as~\cite{DBLP:conf/nips/BrownMRSKDNSSAA20} and~\cite{DBLP:conf/emnlp/ShinRLWS20}. Recently, more studies have focused on soft prompts. \cite{DBLP:conf/acl/LiL20} optimizes the prefix vectors
to solve the text generation task. \cite{DBLP:conf/emnlp/LesterAC21} leverages a trainable vector by adding to the left of context, which shows promising performance on text classification and text generation tasks. More recently, \cite{DBLP:conf/acl/LiuJFTDY022} added a soft prompt to each layer to improve the model's performance in fully supervised scenarios. Unlike other methods which directly optimize the soft prompt by task, \cite{DBLP:conf/acl/GuHLH22} optimizes the soft prompt with a series of pre-training tasks to achieve better few-shot performance in downstream tasks.

\subsection{Controllable text generation}

Controllable text generation has proven successful in text generation tasks. \cite{DBLP:conf/acl/LewisDF18} proposed the Top-k search method to improve the relevance of the story to the prompt. To avoid output text degeneration, Nucleus Sampling~\cite{DBLP:conf/iclr/HoltzmanBDFC20} can draw considerably higher quality text out of neural language models. Moreover, to encourage diversity while maintaining coherence in the generated text, \cite{DBLP:journals/corr/abs-2202-06417} proposed a contrastive training objective to calibrate the model's representation space. To capture fine-grained and more expressive control capability, \cite{DBLP:conf/acl/CarlssonOLVNS22} proposed non-residual attention, whose instruction is equally applicable at any time step.

\section{Methods}

\subsection {Data-free knowledge distillation by text-to-text transfer}
In the simplest form of knowledge distillation (KD)~\cite{DBLP:journals/corr/HintonVD15}, the student model $S$ learns knowledge by training on a transfer set with a soft target provided by the teacher model $ T$'s softmax layer. In the classic distillation setting, the transfer set $\dP$ tends to be the same in the training of both teacher and student models, which is assumed to maximize teacher knowledge acquisition.

In our data-free knowledge distillation (DFKD) setting, the original dataset is unavailable, and the specific domain $\dP$ is unknown for the task of the student model. However, a general domain corpus $\dQ$ can be converted by $F$ (a learnable generative language model), which allows $F(\dQ) \simeq \dP$. Our final goal is to obtain a suitable training corpus $F(\dQ)$ to distill the student model with data with good performance.

\subsection{The transfer text generator}

In the framework of DFKD, we need an $F$ which takes texts from the general domain and outputs the specific domain samples. We model this function as a conditional generation task where a T5 model~\cite{DBLP:journals/jmlr/RaffelSRLNMZLL20} serves as our generator. Inspired by prefix-tuning~\cite{DBLP:conf/acl/LiL20}, which has utilized prompt-tuning for controllable generation, we repurpose the prompting method to steer the pre-trained language model (PLM) to generate the proper transfer data as a good surrogation of $\dP$. Prompt training is more economical than model-tuning, especially when we have a giant PLM. It also substantially reduces the spending on storage and deployment when we need more than one downstream-task student model.

Unlike the previous work of prompt tuning~\cite{DBLP:conf/emnlp/LesterAC21}, which also adds a sequence of embeddings before our input general domain texts, we construct a hybrid prompt. To be more specific, a strong ``commander'' token as a hard prompt is prepended to the other tokens of soft prompts. To reflect the nature of extracting the most meaningful content from the unsupervised corpus, ``summarize'' is chosen as the first "commander" token of our prompt design to confine the orientation of the generation and not to produce repetitive keywords to the downstream task (\eg~poor/beautiful for sentiment classification task). Formally, let $[Pm_i]$ be the $i$-th prompt token and $\bx$ be the text sampled from the public domain dataset. Our final prompt design is as Eq.~\ref{eq_1}.
\begin{equation} \label{eq_1}
\begin{split}
{\rm{summarize:}}[Pm_{0:l}] \bx.
\end{split}
\end{equation}
An ideal objective to tune the soft prompt would be to minimize distribution divergence between the corpus $\dP$ and $F(\dQ)$. However, in our setting, $\dP$ is unknown, and the exact likelihood of $\dQ$ remains intractable, which means that directly calculating the discrepancy is impossible. Therefore, a good surrogate objective needs to be developed to cover the distillation purpose.

\subsection{Specificity Regulator}

The first factor we consider is to maximize the specificity of the generated samples for distillation. Therefore, for classification tasks, an initiative method is to generate samples that the accurate teacher model can confidently classify into a specific class in the task. This indicates that these samples follow the same distribution as that of $\dP$, which is originally noted by~\cite{DBLP:conf/iccv/ChenW0YLSXX019} in computer vision. In our case, after encoding our generated texts from the domain $F(\dQ)$, each token's distribution of the output sequences decoded by our T5 teacher model should be sharp, which means the teacher is more confident. Formally, this training loss function for specificity is formulated as Eq.~\ref{eq_2}.
\begin{equation} \label{eq_2}
\begin{split}
\lsp= \frac{1}{N}\sum_{i}^{N} \sum_{j}^{L} \lce(\bp_i^j, \bt_i^j),
\end{split}
\end{equation}
where $N$ is the batch size, $L$ is the sequence length, $\lce$ is the cross-entropy loss function. $\bp_i^j$ is the distribution of the $j$-th token of $i$-th sequence given by a softmax layer after the teacher's logits outputs as Eq.~\ref{eq_3}.
\begin{equation} \label{eq_3}
\begin{split}
\bp_i^j = {\rm softmax}({\rm logit}\bs_i^j),
\end{split}
\end{equation}
$\bt_i^j$ is the input id of the $j$-th token of $i$-th sequence predicted as Eq.~\ref{eq_4}.
\begin{equation} \label{eq_4}
\begin{split}
\bt_i^j = {\rm argmax}({\rm logit}\bs_i^j).
\end{split}
\end{equation}
Our intuition is that the loss function $\lsp$ could approximate the final objective function, reducing the distance between the distributions of the domain $\dP$ and $F(\dQ)$. During training, we expect the teacher model to be more confident in our adapted samples generated and produce a sequence of tokens with higher probabilities at each corresponding position.

The above case of classification represents a minimalist task for the text-to-text generative model, where a simple prefix (\eg~sentiment analysis) and a single slot output (\eg~positive or negative) are involved. We also conduct the DFKD in the information extraction case to explore a more general application of our method. In particular, this involves a more complicated schema-based instructor as prefix and multiple slots for filling, whose semantics is also richer than class labels. To tackle this new setting, we extend our idea of task specificity and try to find a better surrogate loss. In particular, we introduce a contrastive loss to regulate the balance of classes between slots. For two non-overlapping spans, we have:
\small
\begin{equation} \label{eq_5}
\begin{split}
F(\bm {r}_{i,j},\bm {r}_{\hat i,\hat j}) = \log \frac{\exp({\cos(\bm r_{i,j}, \bm r_{\hat i,\hat j})} / \tau) }{\sum_{\bm r_{m,n}}^{D_{\bar{l}}} \exp(\cos({\bm r_{i,j},\bm r_{m,n}})) / \tau)},
\end{split}
\end{equation}
\normalsize
where $\bm r_{i,j}$ represents a span from the $i$-th to the $j$-th token, $(\hat{i}, \hat{j})$ are sampled non-overlapping spans with $(i, j)$, $\tau$ is the temperature, and $D_{\bar{l}}$ contists of span instances whose label is not $l$. The task specificity objective for IE task is formulated as Eq.~\ref{eq_6}, \ie
\small
\begin{equation} \label{eq_6}
\begin{split}
\lsp = -\sum_{l}^{L} \sum_{r_{{i},{j}} \in D_l }\frac{1}{N_l - 1}\sum_{r_{\hat{i},\hat{j}}\in D_l}F\left( \bm r_{i,j}, \bm r_{\hat{i},\hat{j}} \right ),
\end{split}
\end{equation}
\normalsize
where $L$ represents the number of entity labels, $N_l$ is the number of instances with label $l$ in a batch, and $D_l$ contains all the instances with label $l$. The contrastive loss pushes the entities predicted at different label's slot away from each other and thus increase the specificity of the generated text to each entity label class of generated. Note that a similar strategy has been proposed before~\cite{Si2022SCLRAISC}, but aiming to solve entity ambiguity in NER.

\subsection{Diversity Regulator}

In the distillation process, we need to generate a specific transfer corpus by sampling, and the diversity of the generated samples is also crucial to the distillation performance. This motivates us to introduce our Diversity Regulator to increase generation diversity both at the level of batch and token.

\textbf{Batch Level}: For classification problems, class imbalance is a particular problem of concern. However, as the teacher network has been trained with a particular dataset $\dP$ for a specific task, it is inevitably affected by some kind of class bias in the dataset. As a result, simply optimizing the prompts with our loss of task specificity may cause the generator only to generate samples that the teacher is more confident to give the class label. In other words, this may guide our transfer text generator to choose the easier class among two or more classes for generation. In this sense, we need to encourage our prompts to generate different classes unbiasedly rather than collapse toward easier classes. Therefore, we introduce a diversity regulator at the level of batches. For the text classification task, there exists a set of a limited number $n$ of classes $\mathcal C = \{\bx_{1:l_1}^1, \bx_{1:l_2}^2, \dots, \bx_{1:l_n}^n\}$. Each class label $\bx$ consists of a sequence of tokens to be decoded by the language model guided by the prefix, where $\bx_{1:l_n}^n$ means that the $n$-th class label is composed of $l_n$ tokens (\eg~the class label `very positive' of SST-5 is composed by three tokens, whose IDs are 182, 1465 and 1, respectively, as tokenized by the pretrained T5 tokenizer.). The probability of each class can be calculated by Eq.~\ref{eq_7}. 
\begin{equation} \label{eq_7}
\begin{split}
P^j = \prod^{k}_{l_i}\bp_k,
\end{split}
\end{equation}
where $\bp_k$ is the probability of the $k$-th token. Given a generated sample, we can calculate a probability vector of all the classes $P = (P^1, P^2, \dots, P^n)$ from the outputs of the teacher model. To give a more balanced class distribution, we regulate the diversity of the teacher model's prediction by the diversity loss function as Eq.~\ref{eq_8}.
\begin{equation} \label{eq_8}
\begin{split}
\ldiversity = & \sum_{j}^{n} (\frac{1}{N}\sum_{i}^{N}P_i^j) * \log_{}{(\frac{1}{N}\sum_{i}^{N}P_i^j)} \\
= & -H_{info}(\frac{1}{N}\sum_{i}^{N}P_i)),
\end{split}
\end{equation}
where $P_i$ is the probability vector given by $T$ of the $i$-th sample, and $P_i^j$ is the probability of the $j$-th class. Our loss of diversity is the opposite value of the information entropy of the labels' frequency distribution in each batch. By introducing the diversity loss, we expect that the frequency of each class label that the teacher model predicts would be $\frac{1}{n}$, while any other combination of tokens that the teacher generates will have a frequency of 0.
The final objective can be formulated as Eq.~\ref{eq_9}.
\begin{equation} \label{eq_9}
\begin{split}
\ltt = \lsp + \ldiversity.
\end{split}
\end{equation}
\textbf{Token Level}: As tuning the transfer text generator requires direct gradient backpropagation through the generated sample, we could not decode our text samples by an argmax operation. Therefore, we follow the work of~\cite{DBLP:conf/emnlp/RashidLGR21}, and use the Gumbel-Softmax distribution~\cite{DBLP:journals/corr/KusnerH16} to generate one-hot samples while keeping the differentiability. Different from~\cite{DBLP:conf/emnlp/RashidLGR21}, which only uses Gumbel-Softmax to allow gradient-based backpropagation, we have introduced it to each time step of generation to control the sampling process in a more fine-grained manner. At each time step of generation, the gumbel-softmax trick allows us to sample in the distribution instead of taking argmax operation to generate texts, which essentially increases the diversity of training samples in multiple training epochs. The same general domain text will be transformed into multiple samples close to each other but different, distributed in a neighborhood. This enhances the ability of the prompt to fit the task data $\dP$ leveraged by the teacher model in training.

Formally, we add a stable level parameter $\sigma$ to have
\begin{equation} \label{eq_10}
\begin{split}
\by = {\rm softmax}(\frac{h+\frac{g}{\sigma}}{\tau }),
\end{split}
\end{equation}
\noindent where $\tau$ is a temperature parameter which we keep at 1, $h$ is the logits returned from our generator, and $g$ are sampled from a Gumbel Distribution. When $\sigma \to 0$, the distribution will become a uniform vector, and generated text will become completely random; when $\sigma \to \infty$, the gumbel-softmax trick will degenerate to the argmax operation, and the sampling process will be identical to greedy search. With this stable level parameter, we can regulate the diversity of utterances by adjusting it.

\subsection{Knowledge Distillation}
After prompt-tuning, our generator and prompts are prepared to serve as the transfer text generator to transform the general domain texts to the task-specific dataset. The generated corpus is denoted as $F(\dQ)$ and is used to distill knowledge from our teacher model to the student model.

Because the output of our teacher is a sequence of length greater than one, we need to calculate the divergence between the output distributions of teacher and student at each time step. Taking a batch of $N$ generated samples, we construct the loss of divergence as Eq.~\ref{eq_11}, \ie
\begin{equation} \label{eq_11}
\begin{split}
\ldivergence = \frac{1}{N}\sum_{i}^{N} \sum_{j}^{L} \mathcal{D}(\bp_S^{ij}, \bp_T^{ij}),
\end{split}
\end{equation}
where $\mathcal{D}$ is the divergence function, $\bp_S^{ij}$ and $\bp_T^{ij}$ are the output distributions of the $i$-th sample at time step $j$ of the student model and teacher model respectively. We also follow the work of PKD~\cite{DBLP:conf/emnlp/SunCGL19} to gain generalization ability from the teacher.

Our distillation process also involves the hidden states features as in Eq.~\ref{eq_12}. We apply this loss function for features generated from both the encoder and decoder. To be emphasized, the normalized hidden states are not the direct outputs of Transformer layers because T5 places the layer normalization before the residual skip connection. Taken together, our final distillation objective function is formulated as Eq.~\ref{eq_13}, \ie,
\begin{equation} \label{eq_13}
\begin{split}
\lkd = \alpha \ldivergence + \beta \lhid.
\end{split}
\end{equation}

\section{Experiment and Result Analyses}



We verify the performance of the transfer text generator $ F $ on text classification and text generation,
respectively. We first introduce the implementation details.
Then, we compare the knowledge distillation performance of the model with the $ F $-generated data to the model with various methods and data settings. Finally, we investigate the effectiveness of each component of our proposed method by ablation study and provide more quantitative analyses.

\subsection{Tasks and Datasets}

To evaluate the performance of each model various downstream tasks, we leverage several benchmark text classification datasets: SST-2, SST-5, CoLA, and a named entity recognition dataset: CoNLL03. Note that the extension to name entity recognition is non-trivial, which requires data generation involving multiple slots and schema fitness. Also, in our data-free setting, the original training data is unavailable in training the student model in our data-free setting. These datasets are described in detail in the appendix. For the OOD (out-of-domain) dataset that are fed as $\dQ$ to the text generator, we use the gerneral domain corpus WikiText-103~\cite{DBLP:conf/iclr/MerityX0S17}.

\begin{table*}
\centering
\small
\begin{tabular}{c|c|cccc}
\toprule
\multicolumn{1}{c}{\multirow{3}{*}{\textbf{Method}}} & \multicolumn{1}{c}{\multirow{3}{*}{\textbf{Data}}} & \multicolumn{4}{c}{\textbf{Task/Metric}} \\ \cmidrule(lr){3-6}
\multicolumn{1}{c}{} &\multicolumn{1}{c}{} & \makecell[c]{\textbf{SST-2} \\ ACC} & \makecell[c]{\textbf{SST-5} \\ ACC} & \makecell[c]{\textbf{CoLA} \\ MCC} & \makecell[c]{\textbf{CoNLL03} \\ F1} \\
\midrule
\multicolumn{1}{c}{Teacher} & \multicolumn{1}{c}{Original} & 94.15 & 55.11 & 55.19 & 91.53 \\
\midrule
\multicolumn{1}{c}{\makecell[c]{Student (Standard FT)}} & \multicolumn{1}{c}{Original} & 91.60 & 53.21 & 42.09 & 91.50 \\
\multicolumn{1}{c}{\makecell[c]{Student (Vanilla KD)}} & \multicolumn{1}{c}{Original} & 93.42 $\pm$ 0.43 & 54.21 $\pm$ 0.20 & 44.21 $\pm$ 0.78 & 91.25 $\pm$ 0.18 \\
\midrule
\multicolumn{1}{c}{\makecell[c]{Student (OOD KD)}} & \multicolumn{1}{c}{Wiki} & $87.88 \pm 0.26$ & $44.07 \pm 0.65$ & 22.73 $\pm$ 1.53 & 83.38 $\pm$ 0.22\\
\multicolumn{1}{c}{\makecell[c]{Student (DFKD-T$^3$)}} & \multicolumn{1}{c}{Wiki} & \bm{$91.84 \pm 0.18$} & \bm{$50.09 \pm 1.24$} & \bm{$35.66 \pm 0.80$} & \bm{ $84.65 \pm 0.54 $} \\
\bottomrule

\end{tabular}
\caption{Results of different methods on SST-2, SST-5, COLA, and CoNLL03 datasets. The details of teacher and student models are described in the main text. The first two student models are obtained with the original training data available, which sets an upper bound for the student's performance. In contrast, distillation using OOD data sets a lower bound in the DKFD setting. The standard deviation is calculated with five different random seeds.}
\label{tab:experiments}
\end{table*}

\subsection{Implementation Detail}

\textbf{Text-to-text generator tuning}: We first train the hybrid prompt of following Eq.~\ref{eq_9} and use the trained prompt for data generation in the distillation process. The generator network is trained from a pretrained T5-base model and using Adam optimizer. The training process uses LambdaLR Scheduler to change the learning rate. When the step size is less than 500, the learning rate is 8e-4, which then gradually decays to 8e-5 at step 45000. For the contrastive loss in enity extraction, we use a negative sample ratio of 5.

\textbf{Knowledge distillation}: The distillation protocol is detailed below. To bridge the hidden size of the teacher and student model, we randomly initialize a neural network layer (Eq.~\ref{eq_12}). The weights $\alpha$ and $\beta$ are both set to 1. For a fair comparision, we initialize the student model with an unsupervised pretrained model before knowledge distillation in all settings.

\textit{Text Classification}: We choose T5~\cite{DBLP:journals/jmlr/RaffelSRLNMZLL20} as the experimental model. The teacher models are T5-base models finetuned on each dataset with the original training data, and the student model is T5-small. During the training of the text-to-text generator, the batch size is 32*8. During knowledge distillation, the batch size is set to 32*8.

\textit{Information Extraction}: We use UIE~\cite{DBLP:conf/acl/0001LDXLHSW22} to conduct this task, which can generate contents with a given schema for the information extraction task. The teacher model is UIE-large finetuned on CoNLL03, and the student is UIE-base. During the training of text-to-text generator, the batch size is set to 8*8. During knowledge distillation, the batch size is set to 48*8.

\subsection{Main Result}


Table~\ref{tab:experiments} firstly shows the result of each method on the four benchmark tasks. Note that our experiment is conducted on tasks with different difficulties. For example, SST-2 represents a relatively simple case of binary sentiment analysis, where the student model performs almost on par with the teacher model. In contrast, CoLA is a more difficult linguistic acceptability task where both grammar and semantics are considered.

For each dataset, the performance of the teacher model is listed in the first line, while the corresponding student models' performance using standard finetuning is listed below. Then, we also list the results from the vanilla KD setting. When the original training dataset is available, the student model can be improved compared with the standard finetuning setting, which also serves as an upper bound for the distillation process.

To evaluate the performance of the transfer text generator, a direct comparison is to use the OOD data $\dQ$ as the knowledge distillation data. For a fair comparison, the size of OOD data equals our method's settings. According to Table~\ref{tab:experiments}, using OOD data makes it difficult to achieve good results on these tasks, making this set a lower bound and the Student model's result on original data as the upper bound of the tasks' score.

Unlike directly using OOD data, DFKD-T$^{3}$ trains a text-to-text transfer data generator and achieves impressive margins. Specifically, our Method achieves 3.96\%, 6.02\%, 12.93\%, and 1.27\% absolute improvements in score averaged on SST-2, SST-5, CoLA, and CoNLL03, respectively. This performance demonstrates that the adapter $ F $ makes the samples generated by $ F(\dQ) $ approximate the functionality of the original training set $\dP$.

\subsection{Ablation Studies}


This section mainly discusses the effectiveness of different training designs and how they generate synergies. In the classification tasks, Table \ref{tab:ablation-sentiment} compares different combinations of loss functions in the generation process on the model performance. For the specificity regulator, we show that training the text-to-text model to generate samples with sharpened one-hot classification logits as determined by the teacher model dramatically improves the results on DFKD, especially on CoLA, which are relatively harder to distill with the general corpus (column 1). This highlights the importance of class labels for the transfer text generator.

On the other hand, diversity loss, \ie~the negative information entropy of class distribution at batch level (Eq.~\ref{eq_8}), plays a complementary role as it encourages the even distribution of generated samples. This is especially useful for a multi-class setting. For instance, on SST-5, where a more fine-grained sentiment category is required, applying the diversity loss alone is much better than using the specificity loss alone, showing a 5.3\% higher performance in DFKD results. Combining the two loss functions on all the datasets achieves the highest performance. The synergy of the two-loss designs can be explained as follows. While each sample should be optimized to have a sharp distribution on the class assignment, on the batch level, a balance constraint prevents the generator from collapsing to any single "easy class" in the generation process. We also evaluate the effect of token-level sampling diversity in our diversity regulator (Eq.~\ref{eq_10}), whose result is shown in the Appendix.

\begin{table}
\vspace{-2.0em}
\centering
	\begin{tabular}{c|c|c|c|c}
	\toprule
	$\lsp$ & & \checkmark & & \checkmark \\
	$\ldiversity$ & & & \checkmark & \checkmark \\
	\midrule
	\textbf{SST-2} & $87.88 \pm 0.26$ & $ 89.96 \pm 0.52$ & $91.32 \pm 0.33$ & \bm{$91.84 \pm 0.18$} \\
	\textbf{SST-5} & $44.07 \pm 0.65$ & $44.60 \pm 0.41 $ & $49.94\pm 0.77$ & \bm{$50.09 \pm 1.24$} \\
	\textbf{CoLA} & $22.73 \pm 1.37$ & $34.94 \pm 1.22$ & $35.13 \pm 1.34$ & \bm{$35.66 \pm 0.80 $} \\
	\bottomrule
	\end{tabular}
	
	\caption{Results on different loss functions of the proposed data-free learning method. We compare the model performance with different loss functions. The empty block represents the general domain corpus without any additional optimization. The standard deviation is calculated with different random seeds.}
	\label{tab:ablation-sentiment}
\end{table}

In the case of entity extraction, with the semantic prior given by the extraction schema, we observe that the contrastive loss as the specificity regulator provides a sufficient balance across slot distribution by the pushing-and-pull, leaving the diversity loss redundant but not synergistic in this task. Therefore, we remove the diversity loss for entity extraction, and the improvement in Table~\ref{tab:experiments} comes from the contrastive loss.

\begin{wraptable}{r}{0.4\linewidth}
	\vspace{-1.0em}

	\begin{tabular}{l|l}
	\toprule
	\textbf{Commander Token} & \textbf{SST-2 (ACC)} \\
	\hline
	No ``commander'' & 89.85 $\pm$ 0.29 \\
	\hline
	\textbf{summarize} & \textbf{91.84 $\pm$ 0.18} \\
	adapt & 91.06 $\pm$ 0.23 \\
	expand & 90.60 $\pm$ 0.27 \\
	paraphrase & 90.43 $\pm$ 0.40 \\
	\bottomrule
	\end{tabular}
	
	\caption{Commander Token in hybrid prompt}
	\label{tab:commander}
\end{wraptable}
Our method uses a strong ``commander'' token as the first token of our hybrid prompt. This ``commander'' prepended to the text input serves as a priori constraint on the subsequent generated text to be faithful to the original input, which avoids potential mode collapse problems to some extent. We have tested several commander tokens with empirical results in Table~\ref{tab:commander} showing that ``summarize'' has the best result on knowledge distillation. We have also found that removing this prefix leads to a considerable drop in performance (the accuracy on the dev set of SST-2 drops by 2\%).


\subsection{Comparison with other methods}

To benchmark DFKD-T$^{3}$ with other high-performing DFKD methods, we conduct an additional comparison using the SST-2 task. In particular, as the training of our text generator requires the tokenizer of the text-to-text T5 model, the direct comparison is not possible as the previous methods only compress BERT models. Therefore, to conduct this comparison, we use our generated text as the KD data to distill a BERT-base teacher model to BERT mini, following the protocol of~\cite{DBLP:conf/ijcai/MaWF0022}(Table~\ref{tab:comparison}). Note that this comparison is biased to other methods because their optimization is based on the specific teacher model used, while we solely transfer the data to this new setting. Nevertheless, we outperform the best DFKD method~\cite{DBLP:conf/ijcai/MaWF0022}, which uses an external causal language model for generation, and almost achieves the Vanilla KD result.

\begin{table*}
\vspace{-2.0em}
	\begin{floatrow}
		\capbtabbox{
		\begin{tabular}{c|c}
		\toprule
		\textbf{Methods} & \textbf{SST-2 (ACC)} \\
		\midrule
		Teacher & 93.00 \\
		Vanilla KD & 88.30 \\
		\midrule
		Unlabel KD$^{\clubsuit}$ & 84.90 \\
		Unlabel KD+Adv$^{\clubsuit}$ & 85.90 \\
		PromptDFD-Manual$^{\diamondsuit}$ & 86.35 \\
		PromptDFD-RL$^{\diamondsuit}$ & 87.73 \\
		\textbf{Our Method} & \textbf{88.19} \\
		\bottomrule
		\end{tabular}
	}{
		\caption{Results on the dev set of SST-2. The teacher is BERT-base and the student is BERT-mini. Note that ${\clubsuit}$ is cited from~\cite{DBLP:conf/emnlp/RashidLGR21}, and ${\diamondsuit}$ is cited from~\cite{DBLP:conf/ijcai/MaWF0022}.}
		\label{tab:comparison}
	}

	\capbtabbox{
		\begin{tabular}{c|c|c}
		\toprule
		\textbf{Model } & \begin{tabular}[c]{@{}c@{}}\textbf{\# param.}\end{tabular} & \textbf{\textbf{Inf. time (s)}} \\
		\midrule
		\textbf{CLS-Teacher} & \textasciitilde{}220M ($\times 1.0$) & 60.5 ($\times1.0$) \\
		\textbf{CLS-Student} & \textasciitilde{}60M ($\times 3.6$) & 20.1 ($\times3.0$) \\ \midrule
		\textbf{IE-Teacher} & \textasciitilde{}770M ($\times 1.0$) & 4840.5 ($\times1.0$) \\
		\textbf{IE-Student} & \textasciitilde{}220M ($\times 3.5$) & 1760.2 ($\times2.8$) \\
		
		\bottomrule
		\end{tabular}
	}{
		\caption{Parameter size and inference speed of different models on SST-2 and CoNLL03.}
		\label{tab:speed}
	}
\end{floatrow}
\vspace{-1.0em}
\end{table*}

\subsection{Quantitative Analysis}

\begin{figure*}[!h]
\centering
\begin{center}
\includegraphics[width=1.0\linewidth] {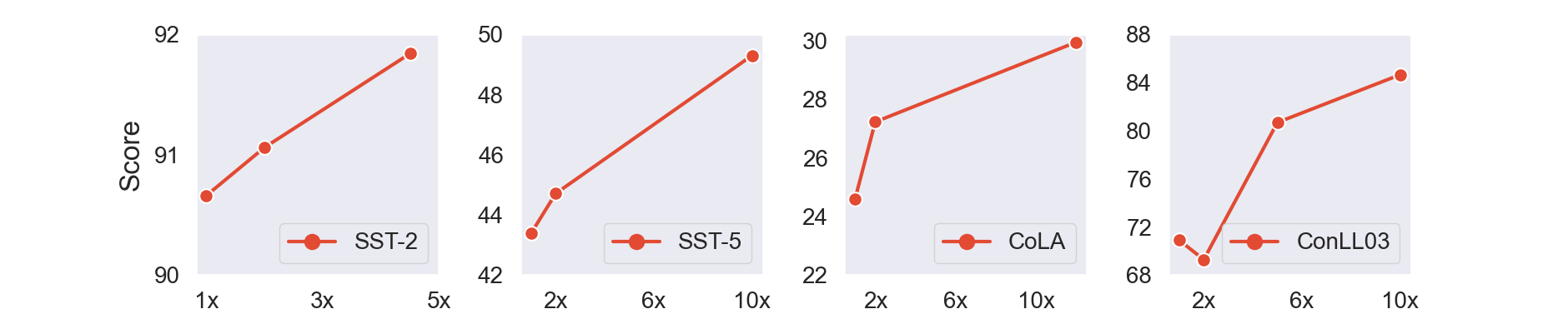}
\end{center}
\caption{Model performance with different data size settings. The horizontal axis of each figure represents the dataset multiplier, and the graph's vertical axis represents the model performance score.}
\label{fig:data size}
\vspace{-1.0em}
\end{figure*}

\textbf{Data scalability}: As the availability of unlabeled corpus is growing tremendously nowadays, here we also investigate how our model scale with different size of corpus size. In Figure \ref{fig:data size}, for each tested task, generated samples with multiple times the amount (up to 10$\times$) of the original training dataset size are selected and used for the knowledge distillation process. Clearly, our results show that increasing the data size of knowledge distillation can significantly improve the performance of models, indicating a great potential for our method to evolve with more data accumulation. 

\textbf{Parameter size and Speed-up}: To demonstrate how the DFKD process can reduce the model parameters and inference time, Table~\ref{tab:speed} shows the model compression ratio and the acceleration rate measured with the Nivida V100 GPU device. Specifically, our classification models are tested on 1K samples from SST-2, and information extraction models are tested on 22K samples from CoNLL03. Taken together with the improved distillation performance, this result indicates an improved performance-efficiency trade-off in the data-free setting.

\section{Conclusion}
We develop a novel framework for DFKD for language models under a text-to-text setting. In particular, we optimize the general domain corpus regarding both specificity and diversity required by distillation by a hybrid prompting strategy. Experiments validate that our method is effective in multiple downstream tasks, outperforms the current SOTA methods, and has excellent data scalability.





\bibliography{custom}
\bibliographystyle{elsarticle-harv}

\clearpage	
\appendix
\section{Dataset details}
This section gives a detailed description of the datasets used in this study.

\textbf{SST-2}: The Stanford Sentiment Treebank is a corpus that allows for sentiment analysis in natural language. The dataset consists of 11k single sentences and each annotated binary sentiment(positive, negative).
The whole dataset is splited to 8.5k training sentences, 1k validation sentences and 1k test sentences. 

\textbf{SST-5}: Same as SST-2's corpus, but it have more sentiment labels: very positive, positive, neutral, negative, very negative.

\textbf{CoLA}: The Corpus of Linguistic Acceptability (CoLA) consists of 10k sentences, the corpus is drawn from books and journals on language theory, and annotated for linguistic acceptability task.

\textbf{CoNLL03}: CoNLL03 is a named entity recognition dataset released as a part of CoNLL-2003 task. For each of the languages there is a training file, a development file, a test file and a large file with unannotated data.
The dataset contains 22k samples and 35k valid entities, which is a representative named entity identification dataset.

\section{Downstream task settings}
\label{sec:appendix-settings}

Our downstream task is based on text-to-text scheme using an encoder-deocder model structure. For classification tasks, we use a prefix corresponding to each task to guide the the model to decode one or two words that corresponds to the target label. For information extraction task, the model generates a specific schema with multiple slots which are filled with entities. The detailed setting of each dataset is listed below. The prefix is for each classification task is underlined. The enenty slot for information extraction is shown in the angle brackets.

\subsection{SST-2}
{
	\setlength{\parindent}{0cm}
	\textbf{Input:} \underline{sst2 sentence:} David Molk is a former american football center. he describes abel dickson as 'a great athlete'</s> </s> </s>
}

{
	\setlength{\parindent}{0cm}
	\textbf{prediction:}  positve</s>
}

{
	\setlength{\parindent}{0cm}
	\textbf{target label:}  1
}

\subsection{SST-5}

{
	\setlength{\parindent}{0cm}
	\textbf{Input:} \underline{sentiment:} soprano, wonderful music, and treatment of the audience. Stunning music from 'e I'm with man' goodman price </s>.
}

{
	\setlength{\parindent}{0cm}
	\textbf{prediction:} very positve</s>
}

{
	\setlength{\parindent}{0cm}
	\textbf{target label:}  4
}

\subsection{CoLA}

{
	\setlength{\parindent}{0cm}
	\textbf{Input:}  \underline{cola sentence:} Doza astralis is one of the most widely cultivated New Zealand trees) </s>
}

{
	\setlength{\parindent}{0cm}
	\textbf{prediction:}  acceptable</s>
}

{
	\setlength{\parindent}{0cm}
	\textbf{target label:}  1
}

\subsection{CoNLL03}

\textbf{Input:} 
<spot> location <spot> miscellaneous <spot> organization <spot> person <asoc> <extra\_id\_2> the review was passed til 1997 but could not be accessed until end of 1997. the commission began to meet mid @-@ January 1998 in the Old Exective Office Bildig and under chairpersonship. olecluiv make</s>

{
	\setlength{\parindent}{0cm}
	\textbf{prediction:}  <extra\_id\_0> <extra\_id\_0> location <extra\_id\_2> Old Exective Office Bildig <extra\_id\_1> <extra\_id\_1>
}

{
	\setlength{\parindent}{0cm}
	\textbf{target label:}  "location": ["Old Exective Office Bildig"]
}

\begin{table*}
	\centering
	\begin{tabular}{c|ccc} 
		\hline
		\multirow{2}{*}{\textbf{Tasks }} & \multicolumn{3}{c}{\textbf{Stable Level }}  \\ 
		\cline{2-4}
		& \textbf{1}   & \textbf{5}   & \textbf{10}   \\ 
		\hline
		\textbf{SST-2}                   & $91.84 \pm 0.18$ & $90.07 \pm 0.79$ & $91.24 \pm 0.63$  \\
		\textbf{SST-5}                   & $50.09 \pm 1.24$ & $48.01 \pm 0.09$ & $48.37 \pm 0.09$  \\
		\textbf{CoLA}                    & $35.66 \pm 0.80$ & $29.21 \pm 0.21$ & $27.56 \pm 0.19$  \\
		\textbf{CoNLL03}                 & $84.65 \pm 0.54$ & $84.47 \pm 0.28$ & $84.04 \pm 0.66$  \\
		\hline
	\end{tabular}
	\caption{Performance on SST-2, SST-5, CoLA, and CoNLL03 with a different stable level of token decoding at inference time. Note that in the prompt tuning processing, our default stable level is set to 1. The standard deviation is calculated with different random seeds.}
	\label{tab:ablation-sl}
\end{table*}

\section{Token-level diversity regulator}

We also evaluate the effect of token-level sampling diversity in our diversity regulator (Eq.~\ref{eq_10}). In Table~\ref{tab:ablation-sl}, by changing the stable level parameter $\sigma$ during text generation for both tasks, we compare the performance of DFKD with decoding diversity. Obviously, the model performance gradually decreases as the stable level of the generation process increases, which highlights a simple yet effective way to boost the DFKD performance.

\section{Data examples by text-to-text transfer}
\label{sec:appendix-synthetic}

In this section, we show some examples for sentiment analysis, linguistic acceptability and information extraction tasks generated by our framework.
We include a few examples of original inputs and its transferred counterpart. An OOD text input us a sentence sampled from wikitext corpus and the generated output is produced by our transfer text generator. We can see the generated senetence contains more related and focused semantics.



\subsection{SST-2}

{
	\setlength{\parindent}{0cm}
	\textbf{OOD Input:} David Molk (born December 15 , 1988 ) is a former American football center. Molk previously played college football at the same position for University of Michigan, where he was a
}

{
	\setlength{\parindent}{0cm}
	\textbf{Generated Output:} David Molk is a former american football center. he describes abel dickson as '\zheyuan{a great athlete}'</s> </s> </s>
}

\subsection{SST-5}

{
	\setlength{\parindent}{0cm}
	\textbf{Original Input:} The Experiment debuted atop the New Zealand Albums Chart on 5 April 2010 , replacing Lady Gaga 's The Fame Monster . In its second charting week.
}

{
	\setlength{\parindent}{0cm}
	\textbf{Generated Output:} soprano, \zheyuan{wonderful} music, and treatment of the audience. \zheyuan{Stunning} music from 'e I'm with man' goodman price </s>.
}

\subsection{CoLA}

{
	\setlength{\parindent}{0cm}
	\textbf{Original Input:} Cordyline australis is one of the most widely cultivated New Zealand native trees , very popular as an ornamental tree in Europe , Great Britain and
}

{
	\setlength{\parindent}{0cm}
	\textbf{Generated Output:} Doza astralis is one of the most widely cultivated New Zealand trees) </s>
}

\subsection{CoNLL03}

\textbf{Original Input:} The second review was to be codcted by a outside commission, althogh it too lawmaers til 1997 to agree po the commission's membership. The commission evetally began to meet in mid @-@ January 1998 in the Old Exective Office Bildig and under the chairmaship of the once

{
	\setlength{\parindent}{0cm}
	\textbf{Generated Output:} the review was passed til 1997 but could not be accessed until end of 1997. the commission began to meet mid @-@ January 1998 in the Old Exective Office Bildig and under chairpersonship. olecluiv make</s>
}

\end{document}